\pdfoutput=1
\PassOptionsToPackage{table, prologue}{xcolor}
\documentclass[11pt]{article}

\usepackage[]{acl}

\usepackage{times}
\usepackage{latexsym}

\usepackage{comment}

\usepackage{microtype}
\usepackage{amsmath,amssymb}
\usepackage{comment}
\usepackage{adjustbox}
	
\usepackage{color, colortbl}

\usepackage{xspace}
\usepackage{booktabs}
\usepackage{multicol}
\usepackage{amsmath}
\usepackage{multirow}
\usepackage{graphicx}
\usepackage{enumitem}
\usepackage{mdwlist}
\usepackage{url}
\usepackage{verbatim}

\usepackage[T1]{fontenc}
\usepackage[utf8]{inputenc}

\usepackage{microtype}
\usepackage{url}

\definecolor{lightgray}{gray}{0.9}
\title{When Do Annotator Demographics Matter? Measuring the Influence of Annotator Demographics with the \textsc{Popquorn} Dataset}

\author{Jiaxin Pei \\
  School of Information \\
  University of Michigan \\
  \texttt{pedropei@umich.edu} \\\And
  David Jurgens \\
  School of Information \\
  University of Michigan \\
  \texttt{jurgens@umich.edu} \\
  }

\newcommand{\sref}[1]{\S\ref{#1}}
\newcommand{\fref}[1]{Figure~\ref{#1}}
\newcommand{\tref}[1]{Table~\ref{#1}}

\newcommand\DATASET{\textsc{Popquorn}\xspace}

\begin{document}
\maketitle

\begin{abstract}

Annotators are not fungible. Their demographics, life experiences, and backgrounds all contribute to how they label data. However, NLP has only recently considered how annotator identity might influence their decisions. Here, we present \DATASET (the \textbf{Po}tato-\textbf{P}rolific dataset for \textbf{Qu}estion-Answering, \textbf{O}ffensiveness, text \textbf{R}ewriting and politeness rating with demographic \textbf{N}uance). \DATASET contains 45,000 annotations from 1,484 annotators, drawn from a representative sample regarding sex, age, and race as the US population. Through a series of analyses, we show that annotators' background plays a significant role in their judgments. Further, our work shows that  backgrounds  not previously considered in NLP (e.g., education), are meaningful and should be considered. Our study suggests that understanding the background of annotators and collecting labels from a demographically balanced pool of crowd workers is important to reduce the bias of datasets. The dataset, annotator background, and annotation interface are available at \url{https://github.com/Jiaxin-Pei/potato-prolific-dataset}.

\end{abstract}

\section{Introduction}
Supervised machine learning relies heavily on datasets with high-quality annotations and data labeling has long been an integral part of the machine learning pipeline \citep{roh2019survey}. While recent large language models show promising performances on many zero-shot and few-shot NLP tasks \citep{bang2023multitask}, reinforcement learning with human feedback (RLHF), the core technology behind these models also heavily relies on large-scale and high-quality human annotations \citep{ouyang2022training, stiennon2020learning}. Therefore, how to curate high-quality labeled datasets is one of the most important questions for both academia and industry. 

Crowdsourcing is actively used as one of the major approaches to collect human labels for various NLP and ML tasks. Early studies on crowdsourcing NLP datasets suggest that crowd workers are able to generate high-quality labels even for relatively difficult tasks and with relatively low costs \citep{snow2008cheap}. However, other studies also suggest that collecting high-quality annotations from crowdsourcing platforms is challenging and requires rounds of iterations to create a reliable annotation pipeline \citep{zha2023data}. 

\begin{table*}[t!]
\centering
\resizebox{2\columnwidth}{!}{%
\begin{tabular}{l|p{4cm}|l|c|c|c|c}
\hline
Task               & Description &Data      & Total Annotations & Number of Annotators & Instances & Average Labels per Instance \\ \hline
Offensiveness rating &  Rate comment offensiveness using a 1-5 scale     & Ruddit    & 13,036            & 262                  & 1,500     & 8.7                        \\ \hline
Question Answering & Read a passage and answer a question through highlighting the text & SQuAD & 4,576             & 459                  & 1,000     & 4.6                        \\ \hline
Text rewriting / Style transfer & Read an email and revise it to make it sound more polite     & Enron     & 2,346             & 257                  & 1,429     & 1.6                        \\ \hline
Politeness Rating & Rate the politeness of an email using a 1-5 scale  & Enron     & 25,042            & 506                  & 3,718     & 6.7                        \\ \hline
\midrule
\DATASET{} & &  & 45,000 & 1,484 & 7,647 & -- \\
\hline
\end{tabular}%
}
\caption{\DATASET{} contains 45,000 annotations from 1,484 participants from a representative sample regarding sex, age and race. Each annotator is paid \$12 per hour as suggested by Prolific. \DATASET{} covers four representative NLP tasks.}
\label{tab:annotation_tasks}
\end{table*}

Annotation quality has typically been measured by proxy through inter-annotator agreement (IAA) metrics like Krippendorff's $\alpha$ \citep{krippendorff2011computing} or Cohen's $\kappa$ \citep{kvaalseth1989note}. 
To attain higher IAA, researchers usually conduct pilot studies or rounds of annotator training to attain higher agreement among annotators. While such a method generally works in settings like part of speech tagging, the use of IAA as a proxy for quality implicitly assumes that the task has real ground truth and disagreements are mistakes. However, annotations for subjective tasks presents a far more challenging setting \citep{sandri2023don}; and as NLP and ML models are more frequently used in social settings where single true answer may not naturally exists, using IAA as the single metric for data quality can be problematic or can even create social harm. For example, \citet{sap2021annotators} studies how annotators' identity and prior belief affect their ratings on language toxicity and found significant gender and race differences in rating toxic language. Other studies also suggest that disagreement in annotations can also be due to the inherent contextual ambiguity \citep{jurgens2013embracing,poesio2019crowdsourced,pavlick2019inherent} which can also be leveraged to improve the model performances \citep{uma2021learning}.

Despite multiple studies on annotator background and disagreement \citep{wan2023everyone, orlikowski2023ecological}, a systematic study on how annotator background influences different types of labeling tasks is still missing in the current literature. To address this gap, in this study, we present \DATASET (the \textbf{Po}tato-\textbf{P}rolific dataset for \textbf{Qu}estion-Answering, \textbf{O}ffensiveness, text \textbf{R}ewriting and politeness rating with demographic \textbf{N}uance) a large dataset labeled by a US-population representative sample of annotators. \DATASET contains 45,000 annotations for four diverse NLP tasks: offensiveness detection (classification/regression), questions answering (span identification), politeness style transfer (language generation) and politeness rating (classification/regression). All four tasks are annotated with a total of 1,484 annotators sampled from a representative sample regarding sex, age and race as the US population.

Through our analysis, we find that demographic background is significantly associated with people's ratings and performance on all four tasks---even for a more objective task such as reading comprehension. For example, people with higher levels of education perform better on the question-answering task and Black or African American participants tend to rate the same email as more polite and the same comment as more offensive than other racial groups. Our study suggests that demographic-aware annotation is important for various types of NLP tasks.

Overall our study makes the following four contributions. First, we create and release \DATASET{}, a large-scale NLP dataset for four NLP tasks annotated by a representative sample of the US population with respect to sex, age, and race. 
Second, we analyze the annotations by different groups of annotators and found that various demographic backgrounds is significantly associated with people's rating of offensiveness, politeness as well as their performance on reading comprehension.
Third, in comparison with existing annotations from curated workers, we demonstrate that a general sample of Prolific participants can produce high-quality results with minimal filtering, suggesting the platform is a reliable source of quality annotations. All the annotations, annotator background information, and labeling interface are available at \url{https://github.com/Jiaxin-Pei/potato-prolific-dataset}.

\section{Motivation}
Individual and group differences are two of the most fundamental components of social sciences \citep{biggs1978individual}. Social and behavioral sciences exist, in part,  because of systematic human variations: if everyone were to behave in the same way, there would be no need to build theories and models to understand people's behaviors in different settings. As a special form of human task, data labeling is also subject to such a basic rule: different people may have different perceptions of various information and different performances on various language tasks. In this sense, while NLP researchers try to achieve a higher IAA, disagreement is a natural and integral part of any human annotation task \citep{leonardelli2021agreeing}. Existing studies in this direction generally focus on building models that can learn from human disagreement \citep{uma2021learning} and some recent studies start to look at how annotators' identity and prior belief could affect their ratings in offensive language and hate speech \citep{sap2019risk, sap2021annotators}. However, most of the existing studies only focus on selected dimensions of identities (e.g., gender) and on certain tasks (e.g., toxic language detection). 

Our study aims at providing a systematic examination of how annotators' background affect their perception of and performances on various language tasks. On the annotator side, we use a representative sample that matches the sex, age and race distribution of the US population. On the task side, we try to select tasks that are representative of common NLP tasks and with different degrees of difficulty, creativity, and subjectivity. Following this criterion, we selected four NLP tasks: (1) offensiveness detection, which is a relatively subjective task for classification and regression, (2) question answering, which is an objective task for span identification that is argued to test reading comprehension, (3) email rewriting, which requires creativity for a text generation task, and (4) politeness rating, which is also a subjective task for classification and regression.

\section{Task 1: Offensiveness detection}

Abusive or offensive language has been one of the most prominent issues on social media \citep{saha2023rise} and many existing studies tried to build datasets and models to detect offensive language \citep{malmasi2017detecting, yin2021towards}. Despite all the efforts on offensiveness detection, these models and datasets may have their own biases and during the creation of these datasets, annotators may introduce their own biases into their labels \citep{sap2019risk}---possibly marginalizing populations whose views differ from the majority. Indeed, \citet{breitfeller2019finding} show that  it was necessary to model the disparity between ratings from men and women annotators to identify gender-based microaggressions.
However, most of the existing studies do not report the background of the annotators \citep{vidgen2020directions}. \citet{vidgen2020directions} reviewed 63 offensiveness datasets and found that only 12 of them report detailed information about annotators.  

To understand how annotator backgrounds (e.g. age, sex and race) affect their ratings on offensiveness, we re-annotated 1500 comments sampled from the Ruddit dataset \citep{hada2021ruddit} using 262 annotators from a representative sample from prolific.co.\footnote{Prolific provides a service to request a sample of annotators with the same distribution of sex, age, and race as the US population using participants self-reported identities. We note that these demographic categories are based on the US Census questions in order to estimate a balanced sample. } In this section, we introduce the data sampling process, annotation task design, annotation result and then discuss how annotators' background affect their ratings of offensiveness.

\subsection{Data and sampling}

\begin{figure}[t]
\centering
\includegraphics[width=0.45\textwidth]{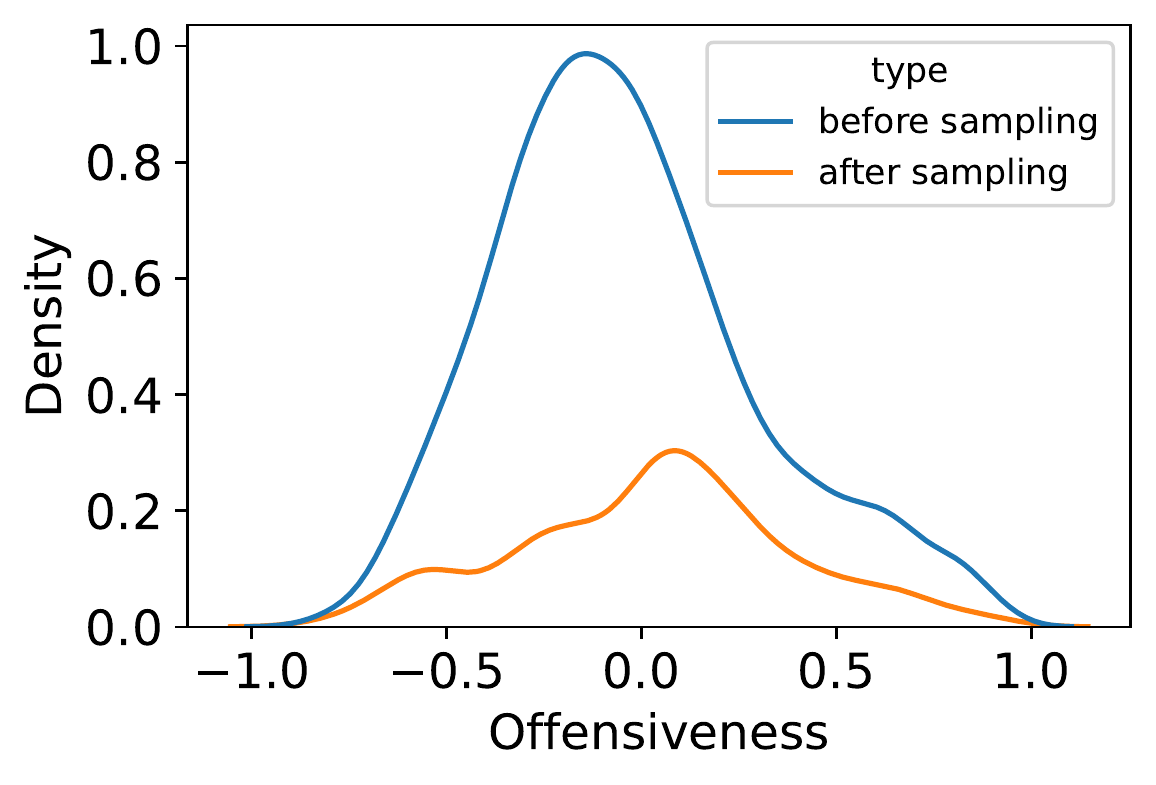}
\caption{The offensiveness score in the Ruddit data before and in our subset after sampling. Positive scores denote offensive text.}

\label{fig:offensiveness-distribution}
\end{figure}

We use the Ruddit dataset \citep{hada2021ruddit} which contains 6,000 Reddit comments annotated using best-worst scaling \citep[BWS;][]{flynn2014best}. Each comment is associated with an offensiveness score ranging from -1 to 1, computed from the BWS ratings. To select the subset we annotate, we remove comments that are shorter than 4 words or longer than 100 words, comments containing URLs as well as quote comments. Such a process led to 5,658 cleaned comments from the Ruddit dataset. We speculated that annotator background might be most influential in borderline cases, i.e., those not extremely offensive or inoffensive. Therefore, we use bucketed sampling based on the offensiveness score and we sample 10\% from (-1, -0.5), 30\% from (-0.5, 0), 50\% from (0, 0.5) and 20\% from (0.5, 1). \fref{fig:offensiveness-distribution} shows the distribution of offensiveness scores before and after the sampling process. Our sampling process produced a subsample of comments with potentially more balanced offensiveness scores.

\subsection{Task design}

Each participant is presented with 50 comments and is asked to rate ``Consider you read the above comment on Reddit, how offensive do you think it is?'' using a 1-5 Likert scale where 1 means ``Not offensive at all'' and 5 means ``Very offensive''. Prior to annotating, each participant is shown an explicit warning about potentially seeing offensive content and has to answer a consent question before any comment is presented. When Prolific provides a demographically-representative sample, some information on the participants is provided. However, to ensure participants consent to have this information shared and reported as they themselves identify, we include a demographic and background screening question after the study is finished. Participants are shown an explanation for why demographic information was being asked for and were allowed to select ``prefer not to disclose'' if they wished.

To validate the annotation procedure, we conducted a pilot study with 8 participants. We used MACE \citep{hovy2013learning} to calculate the annotator competence score and ultimately removed one annotator with a competence score lower than 0.1. The annotators attain moderate IAA (Krippendorff's $\alpha$=0.35), which is on par with existing studies on offensiveness labeling \citep{kang2021style}.  We use \textsc{Potato} \citep{pei2022potato} to set up the annotation website because of its integration with Prolific; Appendix \fref{fig:ui_offensiveness} shows the annotation interface.

\begin{figure}[t]
\centering
\includegraphics[width=0.45\textwidth]{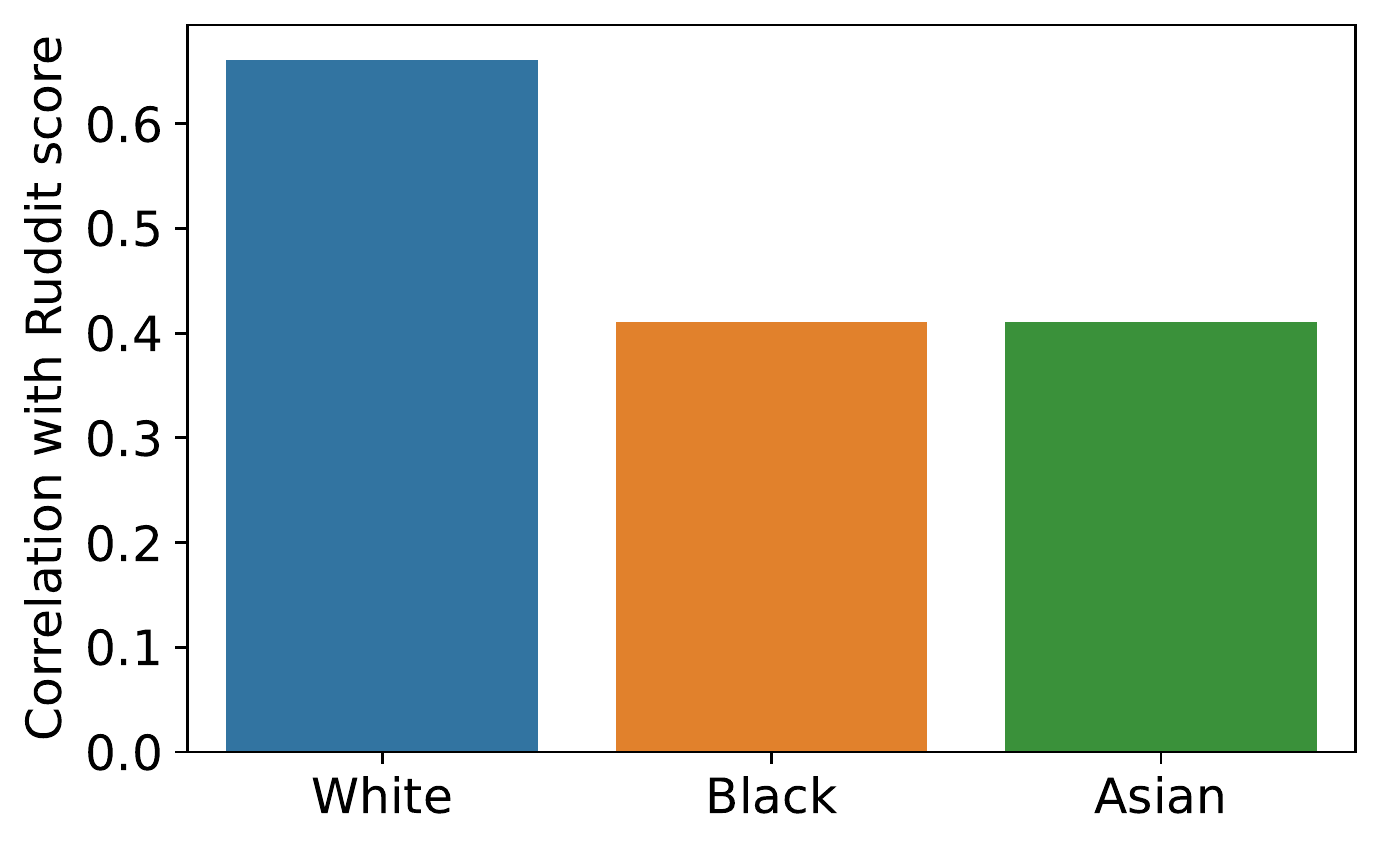}
\caption{Correlation with the original Ruddit offensiveness score by race. Annotations by White participants have the highest correlation with the Ruddit score, while annotations by Asian and Black participants are significantly less correlated.}

\label{fig:offensiveness_corr_by_race}
\end{figure}

\subsection{Annotation result}

The full annotation process collected 13,036 annotations from 262 participants and each comment received 8.7 annotations on average. The medium time of finishing 50 annotations is 13 minutes. Krippendorff's $\alpha$=0.29, showing moderate to low agreement among annotators. However, the overall correlation between the averaged annotations and the original Ruddit score is 0.67, suggesting that, on average, the judgments largely matched those of the original dataset. Participants were highly open to sharing their demographics, with over 95\% filling out the questionnaire.

\begin{table}

\begin{center}\resizebox{.49\textwidth}{!}{%
\begin{tabular}{lrrrr}
\hline
                                  &  Coef. & Std.Err. &      z & P$> |$z$|$ \\
\hline
\textit{intercept}                       &  1.998 &    0.048 & 41.259 &       0.000 \\ 
gender: Non-binary              & {  \cellcolor{red!23}} -0.235 &    0.060 & -3.890 &       0.000 \\ 
gender: Woman                   & {   \cellcolor{red!0}} -0.022 &    0.020 & -1.065 &       0.287 \\
race: Black or African American & {\cellcolor{green!18}}  0.184 &    0.045 &  4.124 &       0.000 \\ 
race: Hispanic or Latino        & {  \cellcolor{red!40}} -0.405 &    0.078 & -5.174 &       0.000 \\ 
race: White                     & {  \cellcolor{red!10}} -0.104 &    0.038 & -2.758 &       0.006 \\
age: 25-29                      & {  \cellcolor{red!19}} -0.185 &    0.043 & -4.268 &       0.000 \\ 
age: 30-34                      & {  \cellcolor{red!17}} -0.165 &    0.041 & -4.071 &       0.000 \\ 
age: 35-39                      & {  \cellcolor{red!14}} -0.142 &    0.040 & -3.525 &       0.000 \\ 
age: 40-44                      & {   \cellcolor{red!0}} -0.037 &    0.043 & -0.860 &       0.390 \\
age: 45-49                      & {  \cellcolor{red!9}}  -0.087 &    0.044 & -1.979 &       0.048 \\
age: 50-54                      & {  \cellcolor{red!14}} -0.141 &    0.046 & -3.077 &       0.002 \\
age: 54-59                      & {  \cellcolor{green!0}} 0.001 &    0.039 &  0.025 &       0.980 \\
age: 60-64                      & { \cellcolor{green!31}} 0.309 &    0.050 &  6.163 &       0.000 \\
age: $>$65                      & { \cellcolor{green!12}} 0.117 &    0.042 &  2.755 &       0.006 \\
education: College degree       & {   \cellcolor{red!0}} -0.015 &    0.023 & -0.660 &       0.509 \\
education: Graduate degree      &  {\cellcolor{green!0}}  0.052 &    0.029 &  1.801 &       0.072 \\
%Group Var                         &  0.435 &    0.022 &        &             &        &         \\
\hline
\end{tabular}}
\end{center}
\caption{Mixed-effect regression results showing the influence of annotator demographics on their offensiveness rating, controlling for the item being rated. Reference categories are Gender: Men, Race: Asian, Age: 18-25, and Education: High school degree.}
\label{tab:offensiveness}
\end{table}

\subsection{Does annotator background affect offensiveness rating?} 
\label{sec:offensiveness-regression}

To understand the influence of annotator background on offensiveness ratings, we ran a linear mixed-effect model to predict the offensiveness rating with gender, age, race, and educational background, controlling each instance as the random effect. By controlling for each instance, we control for differences in the relative levels of offensiveness between instances, which allows us to study deviations from a mean judgment. Categories that are too rare in the data are removed from the regression (e.g. only 1 annotator chooses the ``other'' category for education). 16 annotators are dropped from this process.

\paragraph{Gender} Do men and women have different ratings for offensiveness? Surprisingly, while some existing studies suggest that men and women may have different ratings of toxic language \citep{sap2021annotators}, we found no statistically significant difference between men and women. However, participants with non-binary gender identities tend to rate messages as less offensive  than those identifying as men and women.

\paragraph{Age} People older than 60 tend to perceive higher offensiveness scores than middle-aged participants. It is possible that older people are more sensitive to offensive language and they are less exposed to the language style of Reddit comments. Younger individuals are known to avoid swearing in the presence of older individuals but not among peers \citep[][p.~111]{fagersten2012s} and that younger individuals tend to use stronger swearing \citep{gauthier2017gender}, which supports the idea that inter-generation norms may lead to differences in the perception of toxicity.

\paragraph{Race} We found significant racial differences in offensiveness rating: Black participants tend to rate the same comments with significantly more offensiveness than all the other racial groups. In this sense, classifiers trained on data annotated by White people may systematically underestimate the offensiveness of a comment for Black and Asian people. 

\paragraph{Education} 
No significant differences were found with respect to annotator education, though the relatively small effect for those with graduate degrees does approach significance.

\subsection{Are Ruddit annotations closer to perceptions of people in certain ethnicity groups?} 

We calculate the aggregated score of each racial group and calculate the overall correlation with the Ruddit offensiveness score. As shown in \fref{fig:offensiveness_corr_by_race}, scores by White annotators are highly correlated with the Ruddit annotations (Pearson's $r$=0.66), while the scores by Black, and Asian annotators are only moderately correlated with the Ruddit score (Pearson's $r\approx$0.4), suggesting that the Ruddit annotations are more likely to have been done by White annotators.

\section{Task 2: Question Answering}

Question Answering/Reading comprehension is one of the most fundamental tasks of NLP \citep{rogers2023qa} and SQuAD has been actively used by the research community to evaluate the performance of their models on question answering (QA) as a form of reading comprehension \citep{rajpurkar2016squad,rajpurkar2018know}. To evaluate crowd workers' ability to complete QA tasks and study whether participants' background is associated with different performances, we build the second task as part of the \DATASET{}.

\subsection{Data and sampling}
We use SQuAD 2.0 \citep{rajpurkar2018know} as it also contains unanswerable questions and can pose external challenges to the annotators compared with SQuAD 1.0. In SQuAD 2.0, each passage can contain multiple questions. We sample 1000 unique passages and questions from the SQuAD 2.0 dataset. The final sampled dataset contains 695 questions with correct answers and 305 unanswerable questions. 

\subsection{Annotation task design}
We recruit participants from a US-population representative sample (with respect to sex, age, and ethnicity) on Prolific. Each annotator is assigned with 10 passage and question pairs and is paid \$12 per hour for their participation. At the end of the study, their demographic information is collected through an after-study survey. Besides the question-answering schema, we also ask participants to self-report the difficulty of their questions as task difficulty might be associated with disagreement \citep{uma2021learning}. Appendix \fref{fig:ui_squad} shows the annotation interface for this task.

\subsection{Annotation result} 
4,576 annotations are collected from 459 annotators. Each question received 4.6 annotations on average (similar to the SQuAD data where on average 4.8 answers are collected for each question). We use a similar strategy as \citet{rajpurkar2016squad} to aggregate the answers for each question: choose the majority answer and use the shorter version if there is a tie. We use the evaluation script provided by SQuAD to calculate the token-level precision, recall and F1 score for each answer. The aggregated answers achieve 0.75 F1, 0.72 precision, and 0.79 recall. 

We manually examined a sample of human errors and we found that the crowd workers are mostly able to identify the correct answer but may use a larger span, which leads to higher recall but lower precision. More specifically, we annotated 50 instances where the F1 score is lower than 1 and found that for all these instances, at least one annotator is able to answer it correctly. Moreover, the SQuAD groundtruth is only correct in 12 out of 50 (24\%) instances and for 8 instances (16\%), the crowdworkers are able to identify the correct answer where the SQuAD groundtruth is incorrect. We found 2 out of 50 (4\%) instances that both SQuAD and our crowdworkers didn't answer the question correctly.

\paragraph{What demographic factors influence answer accuracy?}

To study the connection between demographic background and performance on the reading comprehension task, we run a mixed effect model as in \sref{sec:offensiveness-regression} with variables for gender, age, education, and ethnicity as fixed effects and the instance as the random effect.  
Despite the task being largely objective, accuracy at question answering varied relative to annotator background, as shown in \tref{tab:squad}. The largest effects were seen with race and age variation, with a smaller effect for education. While the root causes of this performance disparity cannot be directly tested from our survey, two notable general trends are worth mentioning. 

First, the performance differences mirror known disparities in education and economic opportunities for minorities compared with their White male peers in the US.\footnote{We note that the regression uses the reference category of Asian for race but  the relative differences between groups with factors match with expectations. } Multiple studies have shown how structural forces have led to lower levels of reading abilities by race \citep{dixon2013race,merolla2019structural} and socioeconomic status \citep{merz2020socioeconomic}.

Second, the trend for age  matches known results showing a moderate increase in reading ability with age  \citep{pfost2014individual,locher2020relation}. Further, \citet{locher2020relation} also note that individuals in professions that require reading also have better reading comprehension, which we view as a potential contributor to the performance increase seen from annotators with graduate degrees who are more likely to have such professions.

\begin{table}[t]
\centering
\resizebox{.49\textwidth}{!}{%
\begin{tabular}{lrrrr}
\hline
                                  &  Coef. & Std.Err. &      z & P$> |$z$|$ \\
\hline
\textit{Intercept}                         & {\cellcolor{green!18}}  0.580 &    0.032 & 18.238 &       0.000 \\ 
gender: Non-binary              & {\cellcolor{green!0}}   0.008 &    0.036 &  0.233 &       0.816 \\ 
gender: Woman                   & {\cellcolor{red!10}}  -0.031 &    0.013 & -2.392 &       0.017 \\
race: Black or African American & {\cellcolor{red!23}}  -0.092 &    0.032 & -2.847 &       0.004 \\ 
race: Hispanic or Latino        & {\cellcolor{red!40}}  -0.149 &    0.038 & -3.874 &       0.000 \\ 
race: White                     & {\cellcolor{red!15}}  -0.062 &    0.028 & -2.242 &       0.025 \\
age: 25-29                      & {\cellcolor{green!0}}   0.012 &    0.029 &  0.404 &       0.686 \\ 
age: 30-34                      & {\cellcolor{green!0}}  0.040 &    0.027 &  1.491 &       0.136 \\ 
age: 35-39                      & {\cellcolor{red!0}}  -0.050 &    0.028 & -1.779 &       0.075 \\ 
age: 40-44                      & {\cellcolor{green!18}}  0.072 &    0.028 &  2.567 &       0.010 \\
age: 45-49                      & {\cellcolor{green!20}}  0.079 &    0.027 &  2.903 &       0.004 \\
age: 50-54                      & {\cellcolor{green!29}}  0.116 &    0.029 &  3.938 &       0.000 \\
age: 54-59                      & {\cellcolor{green!18}}  0.072 &    0.027 &  2.697 &       0.007 \\
age: 60-64                      & {\cellcolor{green!0}}   0.002 &    0.027 &  0.060 &       0.952 \\
age: $>$65                      & {\cellcolor{green!0}}   0.008 &    0.026 &  0.311 &       0.756 \\
education: College degree       & {\cellcolor{green!0}}   0.027 &    0.015 &  1.824 &       0.068 \\
education: Graduate degree      & {\cellcolor{green!15}}  0.060 &    0.018 &  3.382 &       0.001 \\
%Group Var                         &  0.041 &    0.010 &        &
\hline
\end{tabular}}
\caption{Mixed-effect regression results showing the influence of annotator demographics on their performance at question answering (as measured by F1 score), controlling for the item being rated. Reference categories are Gender: Men, Race: Asian, Age: 18-25, and Education: High school degree.}
\label{tab:squad}
\end{table}

\begin{figure}[t]
\centering
\includegraphics[width=0.48\textwidth]{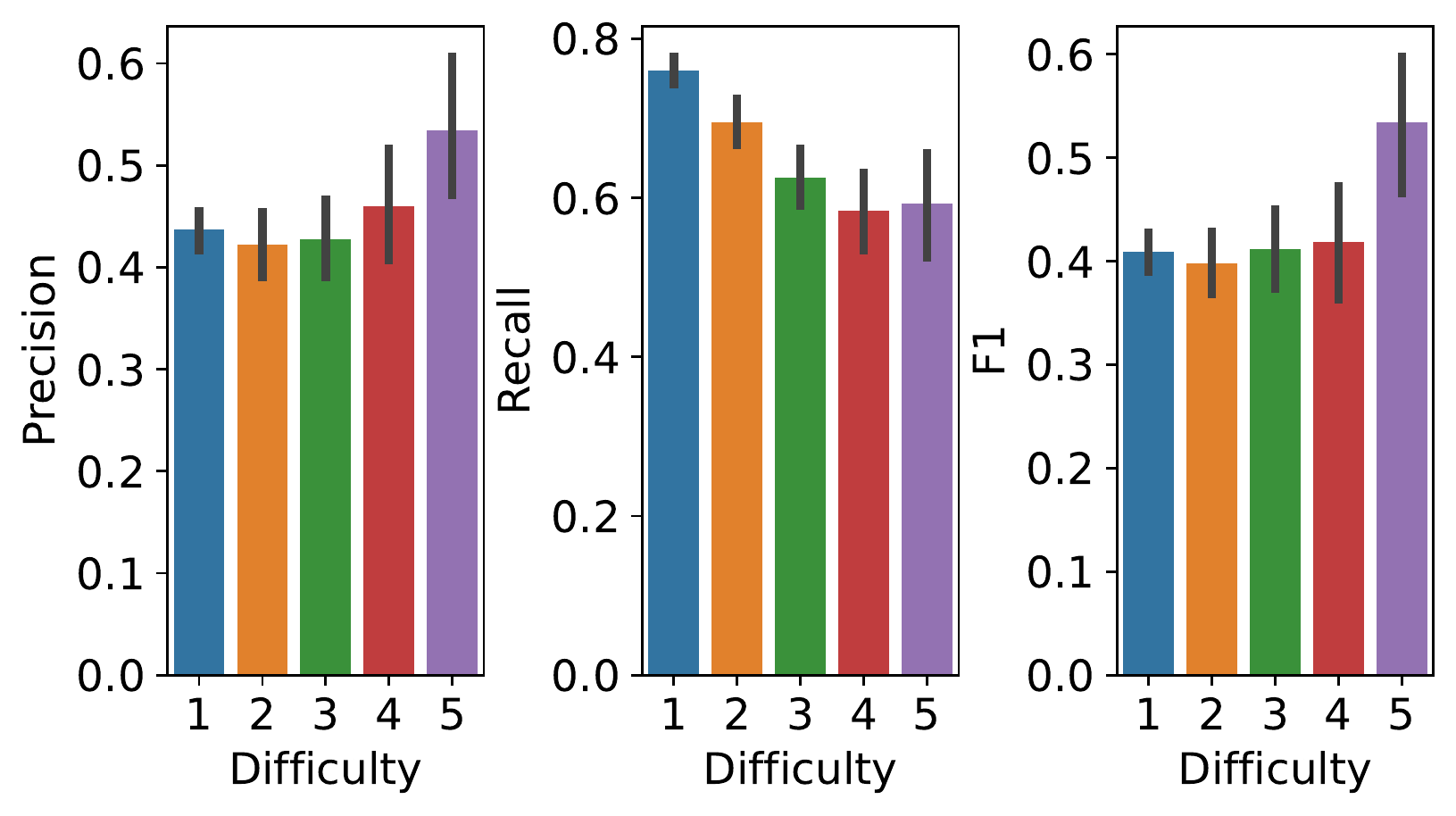}
\caption{Questions rated with lower difficulty are generally associated with higher Recall. However, when people use the highest difficulty score, people generally perform better as measured by precision, suggesting that people tend to be more selective about their answers when they perceive that the task is difficult.}
\label{fig:squad_f1_by_difficulty}
\end{figure}

\paragraph{Is self-reported difficulty associated with participant performance?}  During the study, participants are also asked to rate ``How difficult do you think this question is'' on a 1-5 likert scale where 1 means not difficult at all and 5 means very difficult. \fref{fig:squad_f1_by_difficulty} shows the overall F1, recall and precision score and the difficulty rated by each participant. We found that when people report lower difficulty, their recalls tend to be higher, suggesting that they are better able to identify the potential span of the answer. However, perceived difficulty is also associated with increased precision. Multiple mechanisms might explain this pattern: it is possible that difficult questions require a more specific answer. It is also possible that people may be more cognitively focused to solve the challenge when they perceive the task is more difficult.

\section{Task 3: Politeness rewriting}

Politeness is one of the most prominent social factors in interpersonal communication \citep{brown1987politeness}. The NLP community has built computational models for predicting politeness scores and built models to generate polite text in different settings \citep{danescu2013computational, madaan2020politeness, porayska2004modelling}. However, few resources exist with human-authored examples of pairs of original and style-transferred texts for politeness. Therefore, to test the crowdworker's ability to generate open-domain text for style-transfer tasks, we recruit participants from Prolific to rewrite emails from the Enron dataset as part of \DATASET{}.

\subsection{Data and samples}

\begin{figure}[t]
\centering
\includegraphics[width=0.45\textwidth]{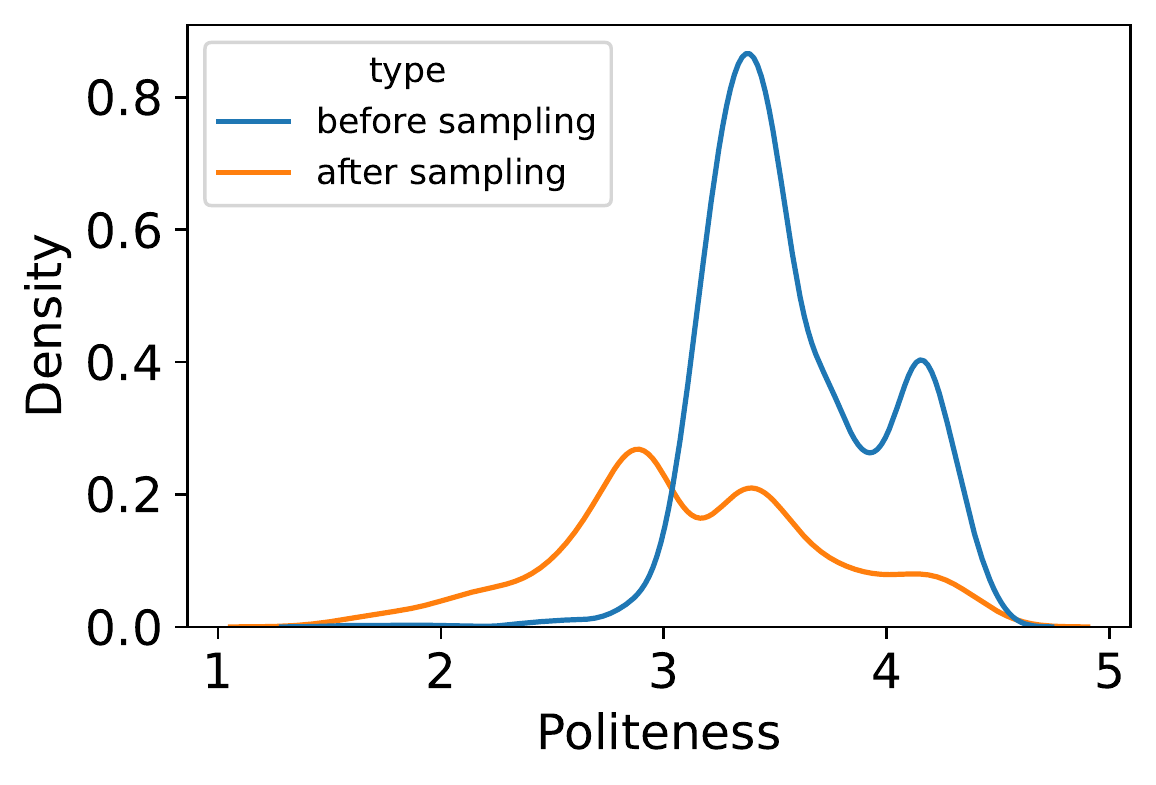}
\caption{The inferred politeness score of emails in the original Enron dataset and in our subset after sampling. Higher scores indicate higher degrees of politeness.}
\label{fig:politeness-distribution}
\end{figure}

We use the Enron email dataset \citep{shetty2004enron} which contains approximately 500,000 emails from senior management executives at the Enron Corporation. We first extract the main body of the emails and then we remove emails that are too long (larger than 100 words), too short (shorter than 8 words), containing URLs, containing more than 10 numbers or were automatically generated by systems. This preprocessing lead to 84,066 remaining emails. We use {\small \texttt{politenessr}} \footnote{The model is accessible at \url{https://github.com/wujunjie1998/Politenessr} and was trained on politeness data from \citet{danescu2013computational} and \citet{wang2018s}.} to infer the politeness score of each email. As most of the emails are relatively polite in the dataset, to draw a more balanced sample for annotation, we use bucket sampling and sample 50\% from (1,3), 40\% from (3,4), and 10\% from (4,5). The final dataset used for annotation contains 1000 emails. \fref{fig:politeness-distribution} shows the distribution of politeness score after bucket sampling. The sampled emails contain more emails with lower inferred politeness scores than the original Enron dataset.

\subsection{Annotation task}

In the annotation task, each annotator is presented with 10 emails and asked to ``rewrite the email to make it sound more polite in a work setting''.   Appendix
\fref{fig:ui_rewriting} shows the annotation interface for this rewriting task. 
We conduct a pilot study with 18 participants to validate the annotation procedure. The pilot study attained 180 annotations for 150 emails and the average editing distance is 102, suggesting that the annotators are making substantial changes to the original message. Politeness of the revised email increases by 0.53 on average when compared with the original emails, suggesting that the revised emails are much more polite.

\begin{figure}[t]
\centering
\includegraphics[width=0.48\textwidth]{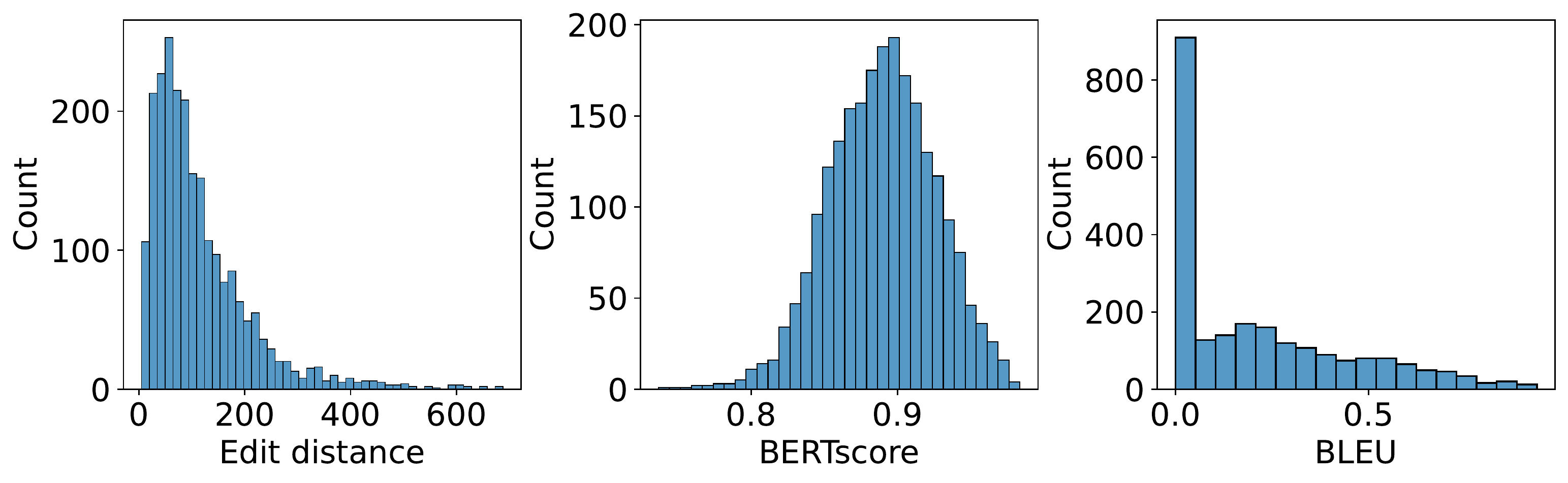}
\caption{Measures comparing the original and revised emails show that the revisions are still very semantically similar (high BERTScore) but the form of the content has been substantially changed (high edit distance and low BLEU score).}
\label{fig:politeness_rewriting_metrics}
\end{figure}

\begin{figure}[t]
\centering
\includegraphics[width=0.48\textwidth]{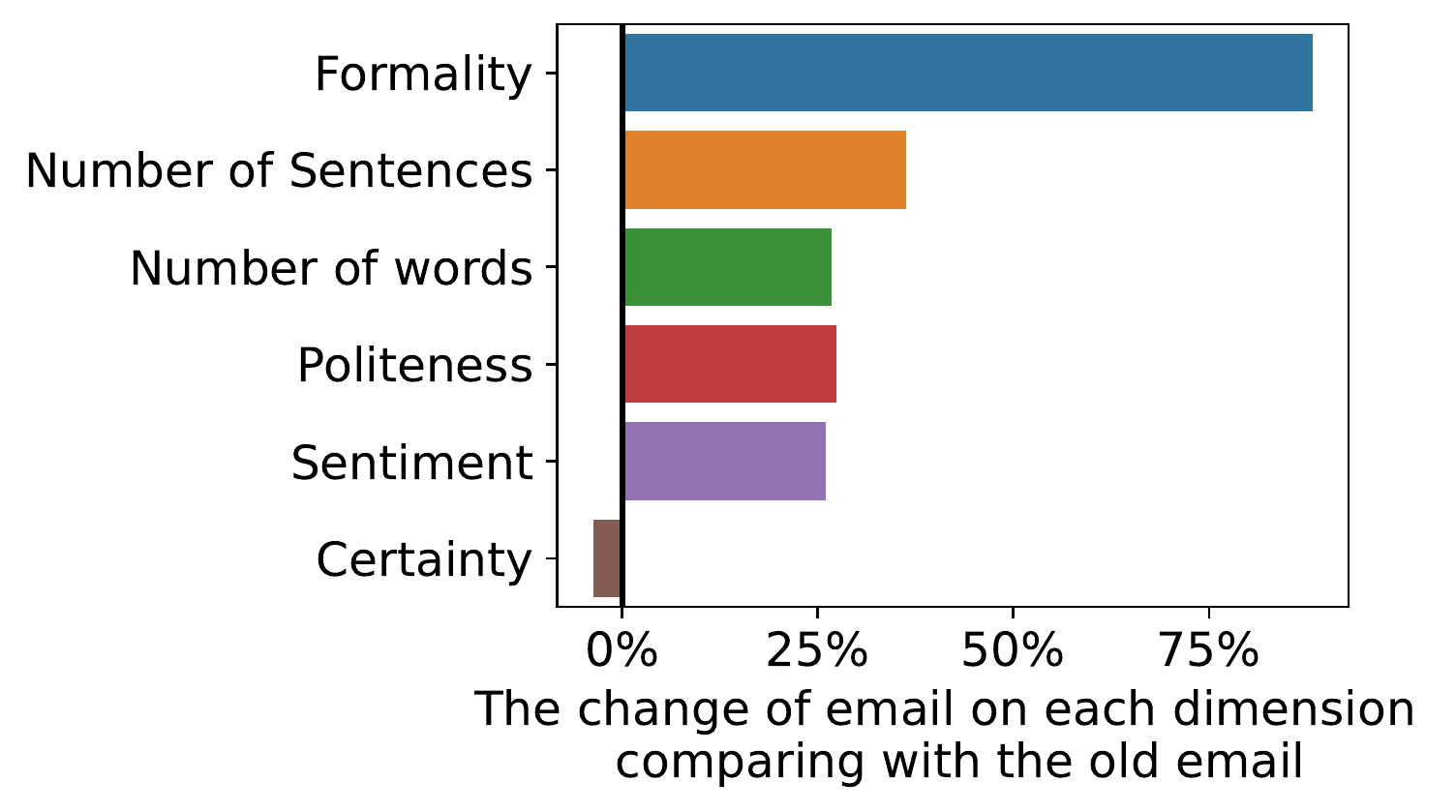}
\caption{The revised emails have 36\% more sentences and 26\% more words than the original emails. Moreover, the revised emails are 88\% more formal, 27\% more polite, 25\% more positive, and 3\% less certain than the old emails, suggesting that the participants are making substantial changes to make the email more polite.}
\label{fig:politeness_change_inferred}
\end{figure}

\begin{figure}[t]
\centering
\includegraphics[width=0.45\textwidth]{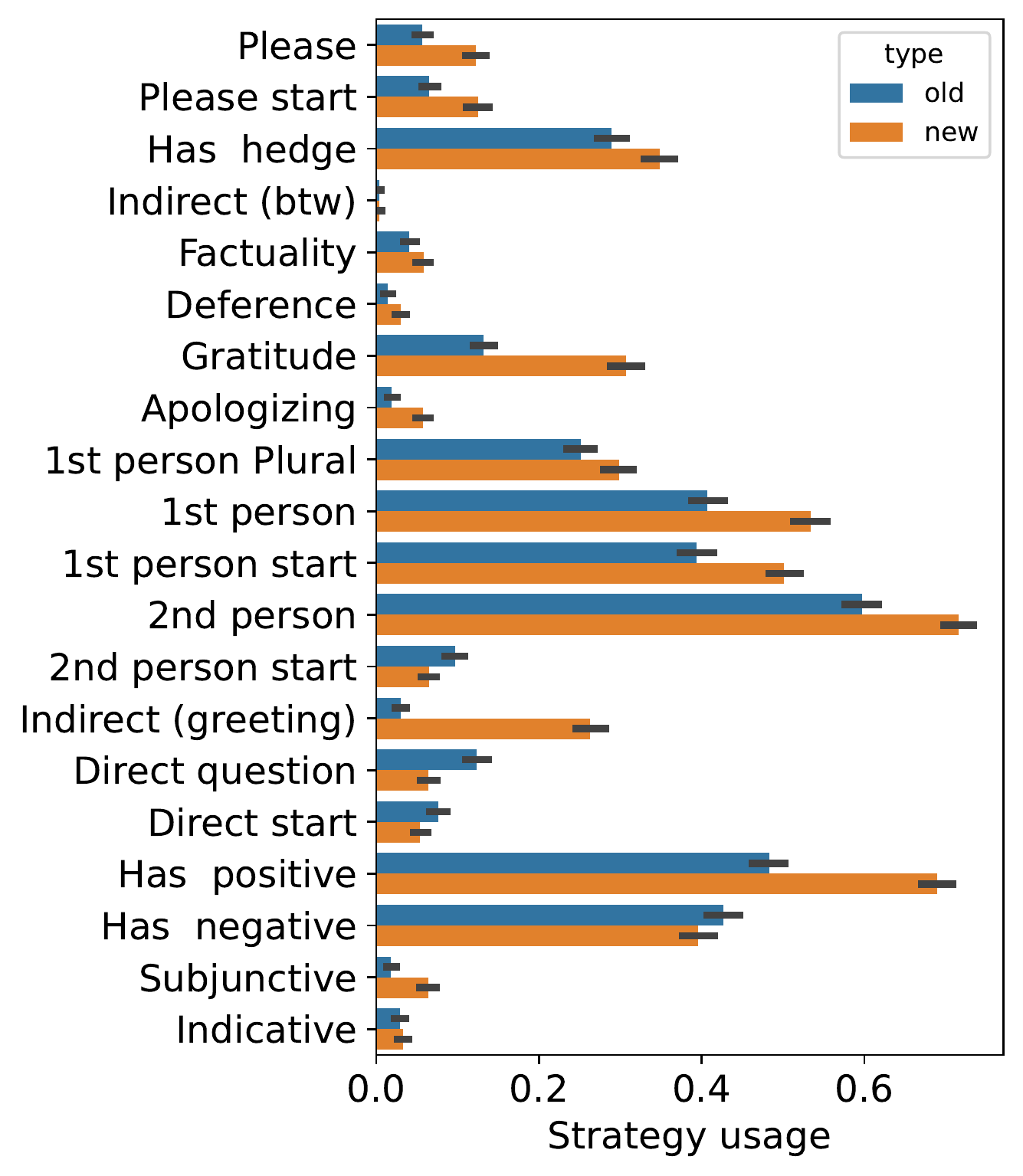}
\caption{Annotators adopt a wide range of politeness strategies.}
\label{fig:politeness_strategies}
\end{figure}

\subsection{Full Annotation Results}

The final politeness rewriting dataset contains 2,346 emails written by 257 participants drawn from a US population representative sample (regarding sex, age, and race). In the final dataset, we remove the revised emails if they are shorter than 7 words or if the edit distance is lower than 5 (79 out of 2376 emails are removed). As shown in \fref{fig:politeness_change_inferred}, the overall politeness increase 27\% compared with the original emails, suggesting that the rewritten emails are significantly more polite than the original ones. The revised emails are more positive\footnote{\url{https://huggingface.co/Seethal/sentiment\_analysis\_generic\_dataset}}, more formal\footnote{\url{https://huggingface.co/s-nlp/roberta-base-formality-ranker}} and less certain\footnote{\url{https://pypi.org/project/certainty-estimator/} \citep{pei2021measuring}} comparing with the original emails.
To achieve these changes, annotators  substantially changed the emails, with an average editing distance of 112; this indicates that changes were mostly not perfunctory, small edits.

Despite these changes to the tone and style of the emails, annotators kept the meaning largely consistent. 
\fref{fig:politeness_rewriting_metrics} shows the distribution of edit distance, BERTscore \citep{zhang2019bertscore} and BLEU, with the latter two being proxies for the interpretation or meaning of the email content. The BERTscore for the emails is generally above 0.8, suggesting that the revised email are able to retain the meaning of the original content. On the other hand, most of the BLEU scores are lower than 0.2, suggesting that the participants are able to make changes to the original content while keeping its meaning. 

Did annotators use a diversity of strategies for increasing politeness---or did they just add ``please'' to every sentence?
To further understand changes annotators made to the original emails, we analyzed the politeness strategies  using ConvoKit \citep{chang2020convokit} and compared the strategies' prevalence in both revised and original emails. As shown in \fref{fig:politeness_strategies}, annotators adopt a wide range of politeness strategies \citep{danescu2013computational}. The usage of ``please'' in a sentence does increase (as expected), and we see a larger increase in strategies such as expressing gratitude, use of positive words, and indirect greetings. Together, this variation suggests that the revisions capture more natural variation in writing and are not artificial revisions driven by task design or speed incentives.

\section{Task 4: Politeness Rating}

To validate the email rewriting results from Task 3, we perform a follow-up participant recruitment to rate the politeness of the original and revised emails. A recent study shows that the nativeness of the annotators affect the IAA of politeness rating \citep{srinivasan2022tydip}.  In this study, we release the full backgrounds of annotators and hope this dataset helps promote future studies on politeness prediction and to understand how people with different backgrounds perceive politeness.

\subsection{Annotation setup}

We use 1,372 emails from the original Enron dataset and 2,346 emails rewritten by the participants. Annotators are asked to rate ``Consider you read this email from a colleague, how polite do you think it is?'' using a 1-5 Likert scale where 1 means ``not polite at all'' and 5 means ``Very polite''. Each annotator is presented with 50 emails in a random order and on average each email is annotated by 6.7 annotators. Appendix \fref{fig:ui_politeness_rating} shows the interface of this annotation task.  We ran one pilot study with 8 annotators and each annotator is presented with 50 emails. The overall Krippendorff's $\alpha$ is 0.43, suggesting moderate inter-annotator agreement and is reasonable for such a subjective task.

\subsection{Full annotations}

Our final politeness rating dataset contains 25,042 annotations from 506 annotators. Each email receives 6.7 annotations on average. The overall Krippendorff's $\alpha$ is 0.43, indicating moderate to low inter-annotator agreement. The overall politeness rating is 2.8 and 3.6 for original and revised emails, suggesting that the revised emails are perceived as more polite than the original emails, which correlates with the previous result.

\subsection{Does annotator background affect politeness rating?} We ran a linear mixed-effect model to predict the politeness rating with gender, age, race, and education, controlling each instance as the random effect, similar to previous setups. \tref{tab:politeness} shows the regression results. 

\paragraph{Gender} We found that women rate messages as less polite, though the effect size is relatively small compared with other demographic dimensions. 

\paragraph{Age} Compared with the youngest segment in our sample (Ages 18-25), all older segments were more likely to give a higher politeness rating. 

\paragraph{Race} We found significant racial differences in politeness rating. 
Relative to Asian peers, Black participants rated messages as more polite, with a small positive effect for White peers. No significant result was seen for annotators identifying as Hispanic or Latino. Given known differences in the cultural perceptions of politeness \citep{troutman2010attitude,brown2015politeness,rodriguez2019you}, these differences suggest systematic variation in the rating that would otherwise be treated as disagreement, rather than valid, culturally-situated judgments.

\paragraph{Educational Background} As shown in \tref{tab:politeness}, participants with more education (a graduate or college degree) tend to rate the same email with less politeness than those with a high school degree. Education is strongly correlated with socioeconomic status and with that status typically comes increased social standing. While multiple works have shown how individuals modify their speech with respect to power/status differences between speaker and recipient \citep[e.g.,][]{brown1987politeness,wang2021price}, we believe our result offers a valuable new insight to how individuals with different status view the \textit{same} message. Our results suggest that higher-status (more educated) individuals are

\begin{table}[tb]
\centering
\resizebox{.49\textwidth}{!}{%
\begin{tabular}{lrrrr}
\hline
                                  &  Coef. & Std.Err. &      z & P$> |$z$|$ \\
\hline
\textit{Intercept}                         & {\cellcolor{green!40}}  3.167 &    0.035 & 89.497 &       0.000 \\ 
gender: Non-binary              & {\cellcolor{red!0}}  -0.048 &    0.042 & -1.149 &       0.250 \\ 
gender: Woman                   & {\cellcolor{red!10}}  -0.042 &    0.014 & -3.116 &       0.002 \\
race: Black or African American & {\cellcolor{green!38}}  0.192 &    0.032 &  6.105 &       0.000 \\ 
race: Hispanic or Latino        & {\cellcolor{green!0}}  0.057 &    0.036 &  1.607 &       0.108 \\ 
race: White                     & {\cellcolor{green!15}}  0.060 &    0.027 &  2.212 &       0.027 \\
age: 25-29                      & {\cellcolor{green!58}}  0.291 &    0.030 &  9.630 &       0.000 \\ 
age: 30-34                      & {\cellcolor{green!19}}  0.078 &    0.028 &  2.764 &       0.006 \\ 
age: 35-39                      & {\cellcolor{green!34}}  0.169 &    0.031 &  5.376 &       0.000 \\ 
age: 40-44                      & {\cellcolor{green!27}}  0.137 &    0.029 &  4.704 &       0.000 \\
age: 45-49                      & {\cellcolor{green!59}}  0.296 &    0.031 &  9.677 &       0.000 \\
age: 50-54                      & {\cellcolor{green!61}}  0.305 &    0.030 & 10.275 &       0.000 \\
age: 54-59                      & {\cellcolor{green!40}}  0.198 &    0.029 &  6.717 &       0.000 \\
age: 60-64                      & {\cellcolor{green!50}}  0.249 &    0.029 &  8.623 &       0.000 \\
age: $>$65                      & {\cellcolor{green!42}}  0.209 &    0.028 &  7.508 &       0.000 \\
education: College degree       & {\cellcolor{red!29}}  -0.145 &    0.015 & -9.394 &       0.000 \\
education: Graduate degree      & {\cellcolor{red!27}}  -0.135 &    0.020 & -6.837 &     0.000 \\
%Group Var & 0.698 & 0.023 & & \
\hline
\end{tabular}}
\caption{Mixed-effect regression results showing the influence of annotator demographics on their politeness ratings, controlling for the item being rated. Reference categories are Gender: Men, Race: Asian, Age: 18-25, and Education: High school degree.}
\label{tab:politeness}
\end{table}

\section{Discussion}

High-quality annotated data has been one of the primary driving factors of NLP and ML. While some studies try to look at improving data quality through analyzing disagreements among annotators, systematic studies of how annotators' background affects crowdsourcing results remain rare. In this paper, we create a new NLP dataset labeled by annotators from a US-representative sample regarding sex, age and race. We re-annotated the Ruddit offensiveness dataset and found that the offensiveness is strongly correlated with annotations by White participants, while the correlation between the Ruddit offensiveness score and annotations by participants from other racial groups are only 0.41, suggesting that the Ruddit dataset might largely reflect the views of White annotators of what content is offensive. As people from other cultures may perceive the same comment with a lower or higher degree of offensiveness, classifiers trained on a dataset annotated by White participants could pose risks for many people. Such an issue becomes increasingly important as both the industry and research community are trying to align the values of LLMs with human beings through instruction tuning.

\section{Conclusion}

Who annotates your data matters. Across four annotation tasks, we show that an annotator's background influences their decisions, across multiple annotation tasks with different degrees of subjectivity.  In more subjective tasks, these differences in decisions are not mistakes but rather valid differences in views. Our results underscore that NLP papers that curate datasets must consider whose voices appear in their datasets, as these ultimately decide whose voice are captured in models trained on the data. Indeed, by comparing our annotations with those from the  existing annotated datasets, we show that the existing annotated dataset might be annotated by a demographically-biased group of annotators. To support work in this modeling demographic-aware and socially-responsible NLP, we release \DATASET with 45,000 annotations on four NLP tasks by nearly 1.5K annotators.

\section*{Acknowledgments}

We thank Prolific for their generous support to this study. We thank Tom Rodenby, Andrew Gordon and Agrima Seth for their helpful feedback. This material is based in part upon work supported by the National Science Foundation under Grant No IIS-2143529.

\section*{Ethical Implications}
Collecting background information about annotators can be sensitive and have ethical implications. In our study, we follow best practices when asking about demographic information \citep{spiel2019better} and always allow participants to choose ``Prefer to not disclose'' and provide external options for them to self-describe identities. Understanding how different groups of people perceive social information in language and perform different tasks is important when NLP models are applied in more and more social applications. We believe that through carefully designed procedures to collect background information of annotators along with the data annotation, we will be able to build better NLP and ML models that could better serve different groups of people and reduce potential social harm.

\bibliography{anthology,custom,references}
\bibliographystyle{acl_natbib}

\clearpage

\appendix

\begin{figure*}[h]
\centering
\includegraphics[width=0.95\textwidth]{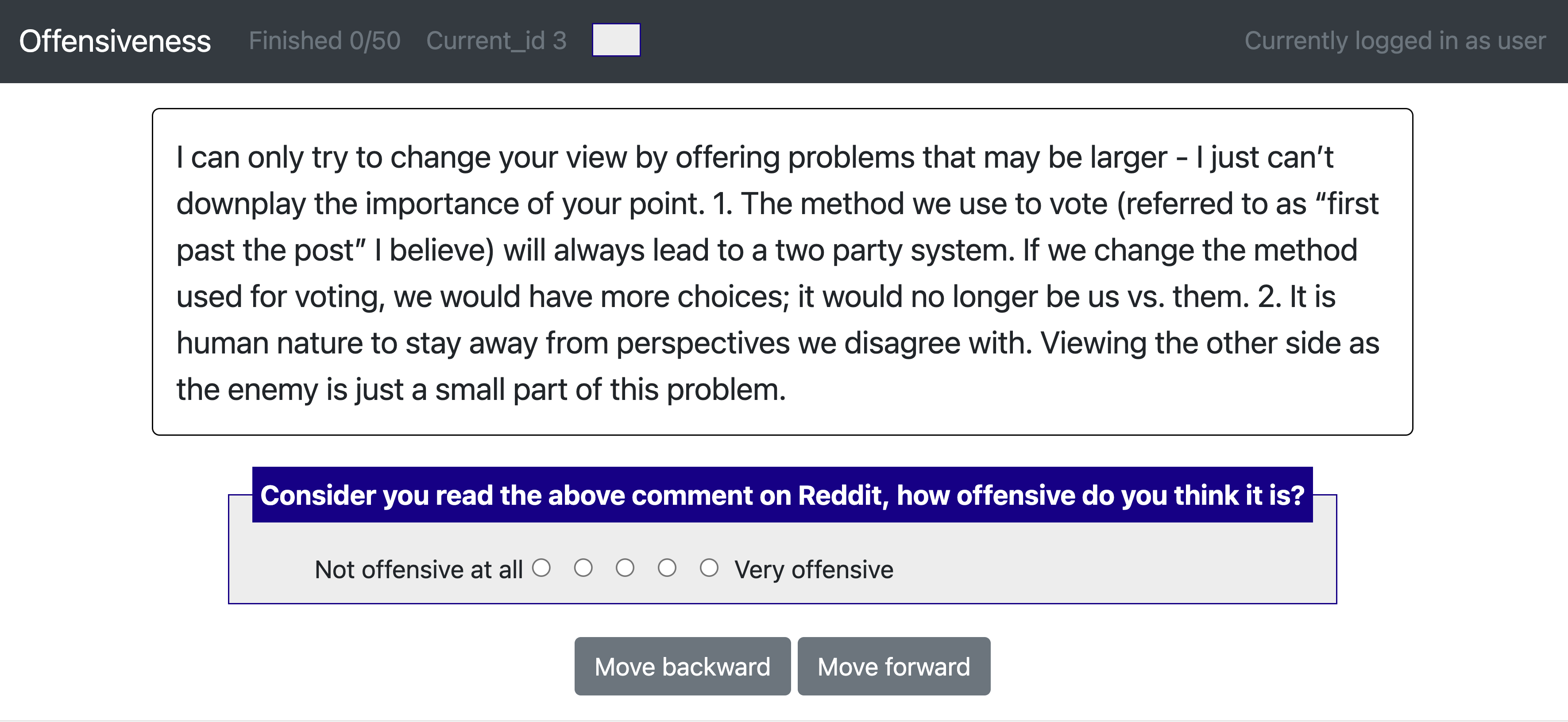}
\caption{Annotation interface for the offensiveness rating task.}
\label{fig:ui_offensiveness}
\end{figure*}

\begin{figure*}[h]
\centering
\includegraphics[width=0.95\textwidth]{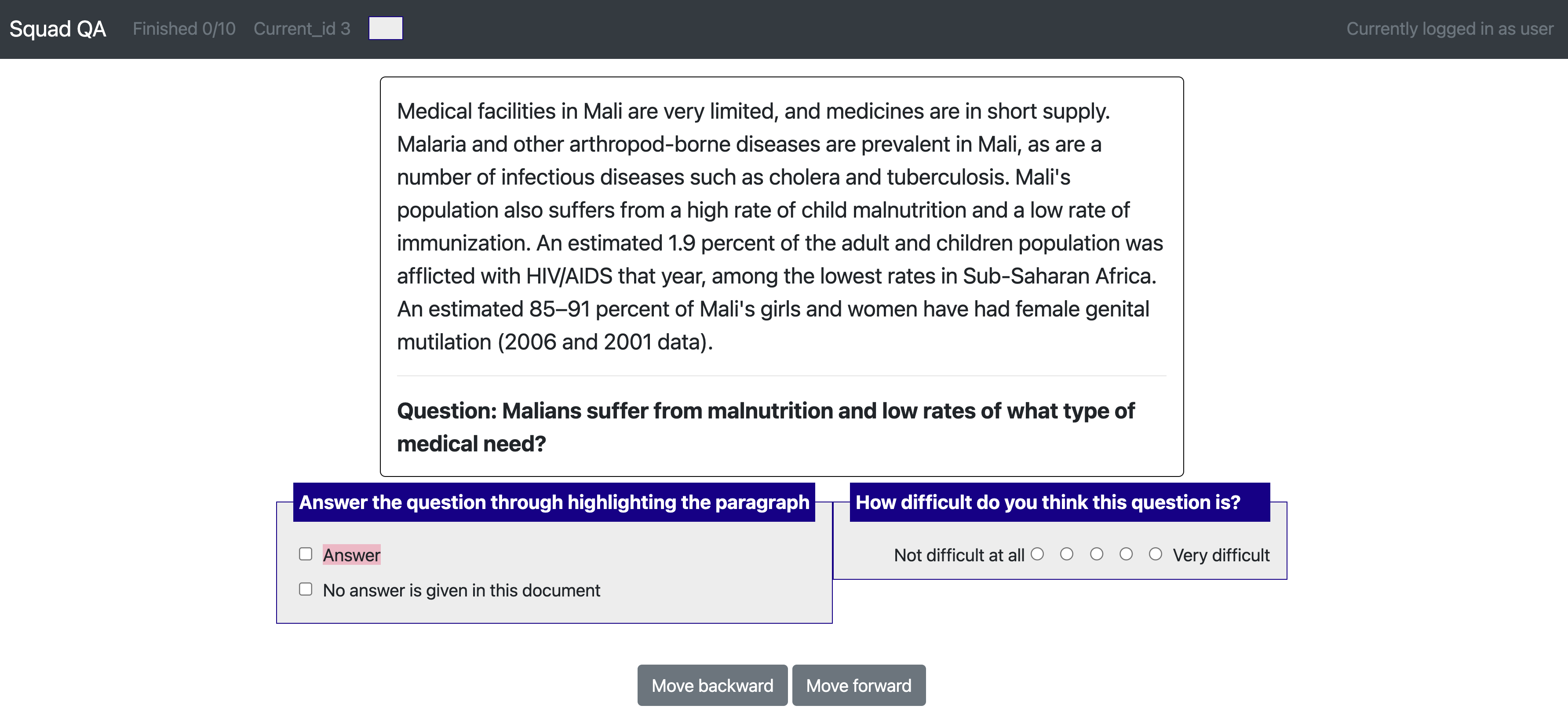}
\caption{Annotation interface for the SQuAD reading comprehension task}
\label{fig:ui_squad}
\end{figure*}

\begin{figure*}[t]
\centering
\includegraphics[width=0.95\textwidth]{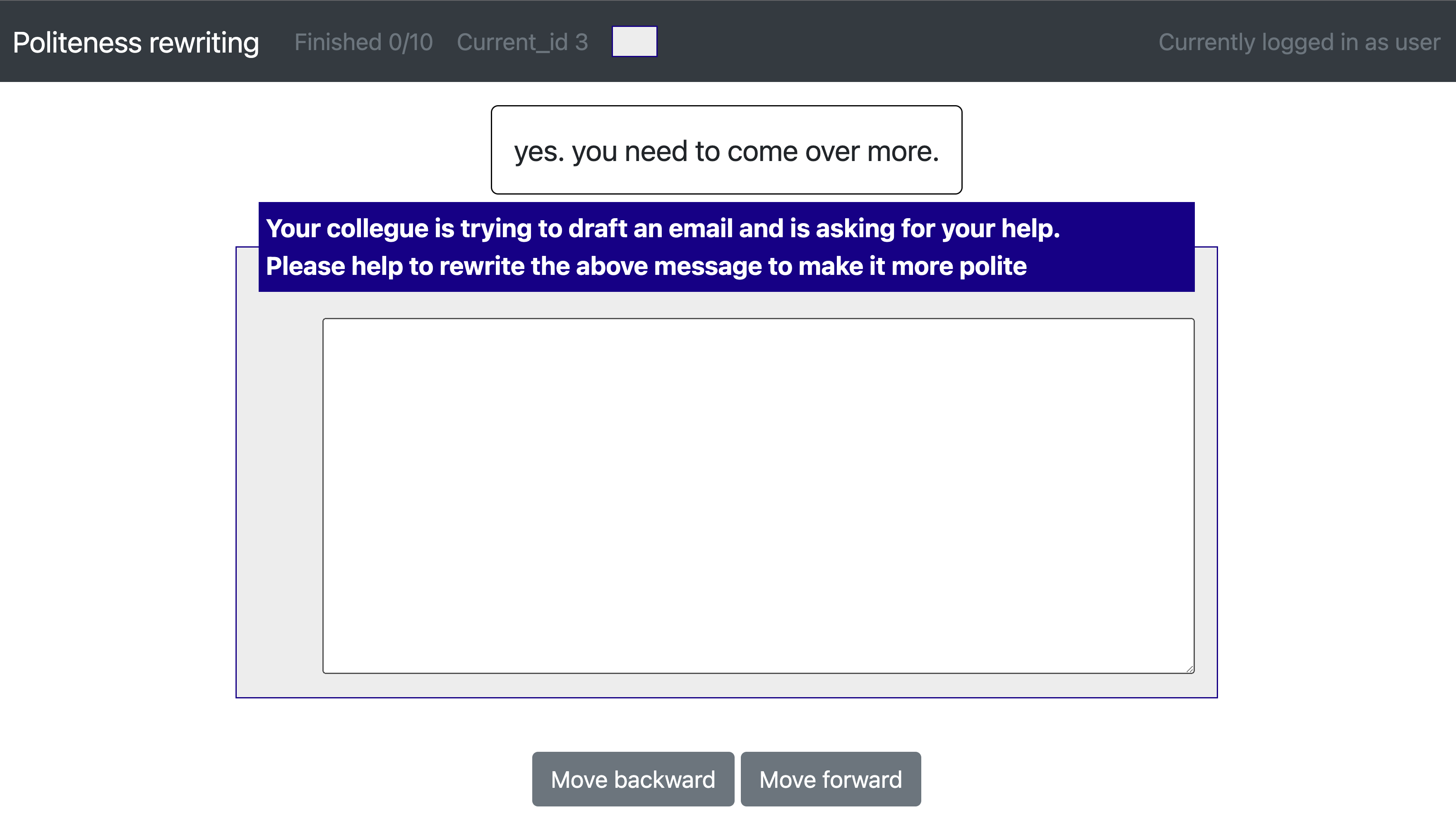}
\caption{Annotation interface for the email rewriting task}
\label{fig:ui_rewriting}
\end{figure*}

\begin{figure*}[t]
\centering
\includegraphics[width=0.95\textwidth]{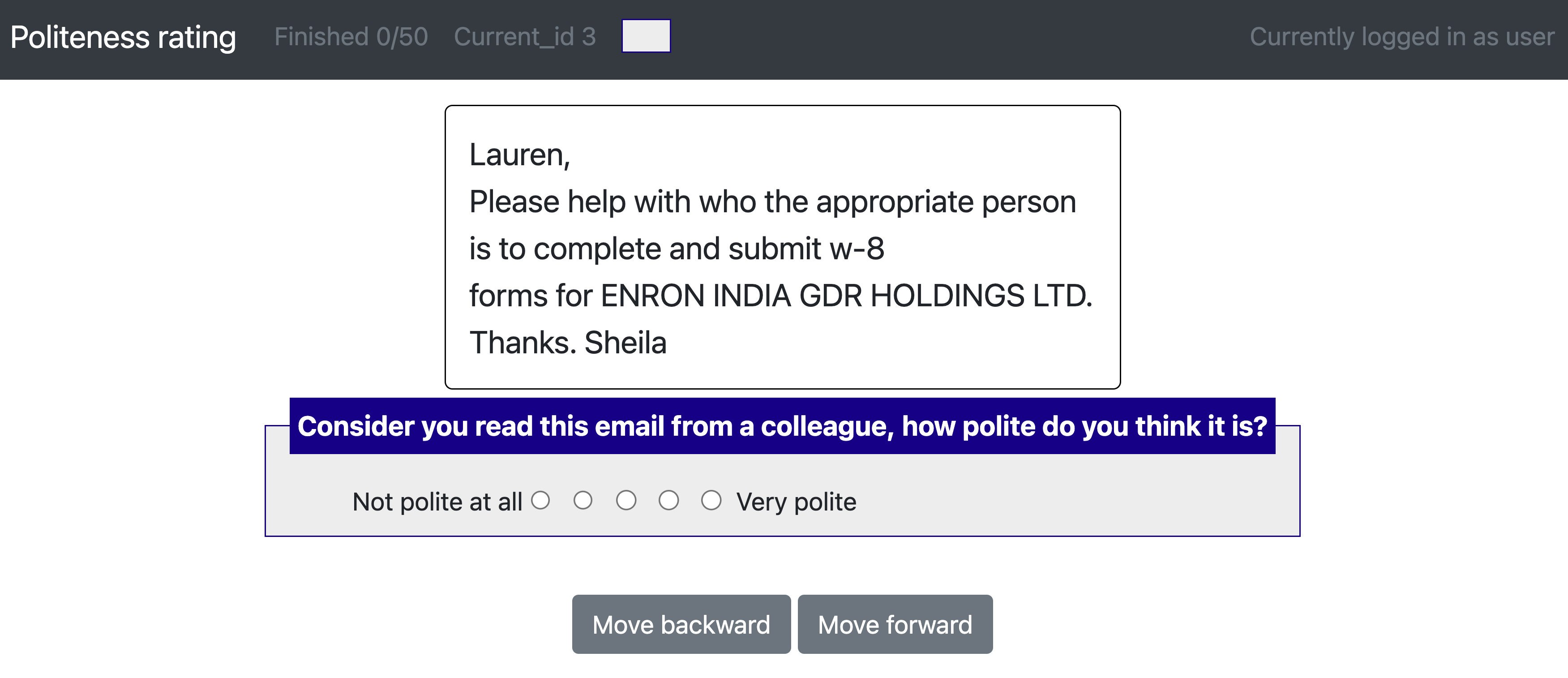}
\caption{Annotation interface for the politness rating task.}
\label{fig:ui_politeness_rating}
\end{figure*}

%\section{Example Appendix}
%\label{sec:appendix}

%This is an appendix.

\end{document}